%% file: ms.tex
\documentclass[letterpaper, 10 pt, conference]{ieeeconf}  

\IEEEoverridecommandlockouts                              

\usepackage{graphics} 
\usepackage{epsfig} 
\usepackage{mathptmx} 
\usepackage{times} 
\usepackage{amsmath} 
\usepackage{amssymb}  
\usepackage{tikz}
\usepackage{array}
\input{Commands}
\usepackage{multirow}

\makeatletter
\newcommand\copyrighttext{%
  \footnotesize \copyright 2018 IEEE. Personal use of this material is permitted.
  Permission from IEEE must be obtained for all other uses, in any current or future
  media, including reprinting/republishing this material for advertising or promotional
  purposes, creating new collective works, for resale or redistribution to servers or
  lists, or reuse of any copyrighted component of this work in other works.
  DOI: 10.1109/ICRA.2018.8460490}

\newcommand\copyrightnotice{%
\begin{tikzpicture}[remember picture,overlay]
\node[anchor=south,yshift=10pt] at (current page.south) {\fbox{\parbox{\dimexpr\textwidth-\fboxsep-\fboxrule\relax}{\copyrighttext}}};
\end{tikzpicture}%
}

\makeatother

\title{\LARGE \bf
Mono-Stixels: Monocular depth reconstruction of dynamic street scenes
}

\author{Fabian Brickwedde$^{1,2}$, Steffen Abraham$^{1}$, Rudolf Mester$^{2}$
\thanks{$^{1}$ Robert Bosch GmbH, Hildesheim, Germany
				{\tt\small firstname.lastname@de.bosch.com}}%
\thanks{$^{2}$ VSI Lab, Goethe University, Frankfurt, Germany
        {\tt\small mester@vsi.cs.uni-frankfurt.de}}%
}

\begin{document}

\maketitle

\thispagestyle{empty}
\pagestyle{empty}
\copyrightnotice

\begin{abstract}
\input{text/Abstract/Abstract}
\end{abstract}

\section{INTRODUCTION}
\input{text/Introduction/Introduction}
\section{RELATED WORK}
\input{text/Introduction/RelatedWork}
\section{METHOD}
\input{text/Method/Method}

\section{EXPERIMENTS}
\input{text/Experiments/Experiments}

\section{CONCLUSIONS}
\input{text/Conclusion/Conclusion}

\addtolength{\textheight}{-6cm}   


\bibliographystyle{bibtex/IEEEtran}
\bibliography{bibtex/IEEEabrv,Literatur/Literatur}





\end{document}

%% file: Commands.tex
\renewcommand{\matrix}[1]{\begin{bmatrix} #1 \end{bmatrix}}
\newcommand\RotText[1]{\rotatebox{90}{\parbox{1.25cm}{\centering#1}}}

\newcommand{\vectorsymbol}[1]{\mathbf{#1}}
\newcommand{\vx}{\vectorsymbol{x}}

\newcommand{\vt}{\vectorsymbol{t}}
\newcommand{\vr}{\vectorsymbol{r}}
\newcommand{\vb}{\vectorsymbol{b}}
\newcommand{\vn}{\vectorsymbol{n}}
\newcommand{\vf}{\vectorsymbol{f}}

\newcommand{\vs}{\vectorsymbol{s}}

\newcommand{\vl}{\vectorsymbol{l}}

\newcommand{\matrixsymbol}[1]{\mathbf{#1}}
\newcommand{\mA}{\matrixsymbol{A}}
\newcommand{\mR}{\matrixsymbol{R}}
\newcommand{\mK}{\matrixsymbol{K}}
\newcommand{\mH}{\matrixsymbol{H}}

\newcommand{\cov}{\mathbf{C}}

\renewcommand{\sup}{\mathbb{G}}
\newcommand{\sky}{\mathbb{S}}

\newcommand{\dobj}{\mathbb{DO}}
\newcommand{\sobj}{\mathbb{SO}}

\newcommand{\Onot}[1]{\mathcal O\left( #1 \right)}

%% file: text/Abstract/Abstract.tex
In this paper we present mono-stixels, a compact environment representation specially designed for dynamic street scenes. Mono-stixels are a novel approach to estimate stixels from a monocular camera sequence instead of the traditionally used stereo depth measurements. \par 
Our approach jointly infers the depth, motion and semantic information of the dynamic scene as a 1D energy minimization problem based on optical flow estimates, pixel-wise semantic segmentation and camera motion. The optical flow of a stixel is described by a homography. By applying the mono-stixel model the degrees of freedom of a stixel-homography are reduced to only up to two degrees of freedom. Furthermore, we exploit a scene model and semantic information to handle moving objects. \par
In our experiments we use the public available DeepFlow for optical flow estimation and FCN8s for the semantic information as inputs and show on the KITTI 2015 dataset that mono-stixels provide a compact and reliable depth reconstruction of both the static and moving parts of the scene. Thereby, mono-stixels overcome the limitation to static scenes of previous structure-from-motion approaches.

%% file: text/Introduction/Introduction.tex
Autonomous vehicles and advanced driver assistance systems need to understand the surrounding environment to identify critical objects, 
parking slots or navigate through the street scene. 
Therefore, a representation of the geometric and semantic layout of the street scene is necessary.

One useful compact medium-level representation for street scenes is the so called stixel world that was introduced by Badino et al. 
  \cite{badino2009stixelworld} 
and extended to a multi-layer stixel world by Pfeiffer and Franke \cite{pfeiffer2011multilayer}. 
The stixel world is defined as a column-wise segmentation of the image in thin sticklike planar elements, the stixels. 
The underlying world model distinguishes three types of stixels: lying ground stixel, upright object stixel and sky stixel at infinite distance. 
Thus, the geometric layout of a stixel can be described by one value for the depth resulting in a quite compact representation. 
Furthermore, Schneider et al. \cite{schneider2016semantic}  introduce the semantic stixels that additionally consist of a semantic label like road, vegetation or vehicle. 
Many constructive works show that this medium-level representation is suitable for many high-level vision tasks 
like object segmentation \cite{erbs2014spider}, object tracking \cite{pfeiffer2011dynamicstixel} or region of interests selection for pedestrian detection \cite{benenson2012fast}. 
Even autonomous driving based on the stixel world seems to be possible as shown by Franke et al. \cite{franke2013making}.

\begin{figure}[t]  
  \centering
	\extracolsep{\fill}
	  \begin{tabular}{@{}c@{} c@{}}
		 \begin{tikzpicture}[]
			\node[style={inner sep = 2, outer sep=0}] (pic) at (0,0) {\includegraphics[width=0.23\textwidth]{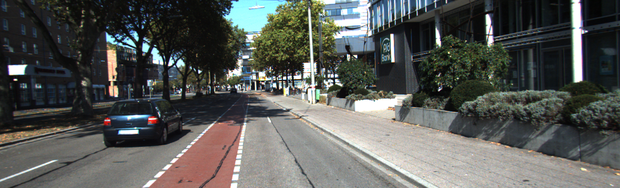}};
			\node[style={outer sep=0}, below left] at (pic.north east) {\small{\textcolor{white}{Image}}};
		  \end{tikzpicture}
		 & \begin{tikzpicture}[]
			\node[style={inner sep = 2, outer sep=0}] (pic) at (0,0) {\includegraphics[width=0.23\textwidth]{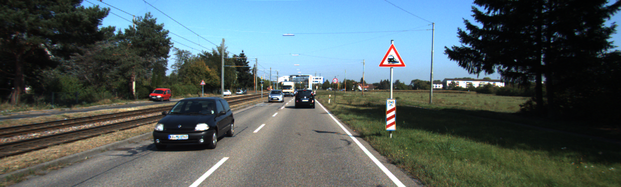}};
			\node[style={outer sep=0}, below left] at (pic.north east) {\small{\textcolor{white}{Image}}};
		  \end{tikzpicture}\\
					 \begin{tikzpicture}[]
			\node[style={inner sep = 2, outer sep=0}] (pic) at (0,0) {\includegraphics[width=0.23\textwidth]{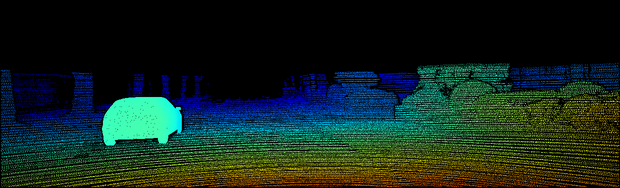}};
			\node[style={outer sep=0}, below left] at (pic.north east) {\small{\textcolor{white}{Ground Truth}}};
		  \end{tikzpicture}
		 & \begin{tikzpicture}[]
			\node[style={inner sep = 2, outer sep=0}] (pic) at (0,0) {\includegraphics[width=0.23\textwidth]{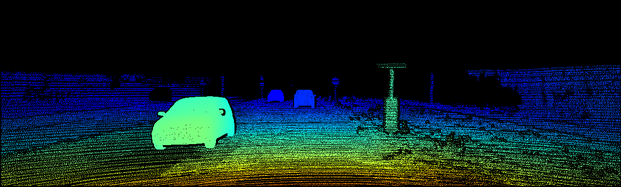}};
			\node[style={outer sep=0}, below left] at (pic.north east) {\small{\textcolor{white}{Ground Truth}}};
		  \end{tikzpicture} \\	
		\begin{tikzpicture}[]
			\node[style={inner sep = 2, outer sep=0}] (pic) at (0,0) {\includegraphics[width=0.23\textwidth]{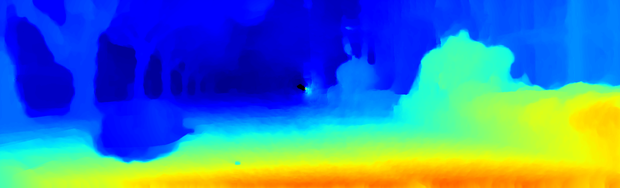}};
			\node[style={outer sep=0}, below left] at (pic.north east) {\small{\textcolor{white}{SFM}}};
		  \end{tikzpicture}
		 & \begin{tikzpicture}[]
			\node[style={inner sep = 2, outer sep=0}] (pic) at (0,0) {\includegraphics[width=0.23\textwidth]{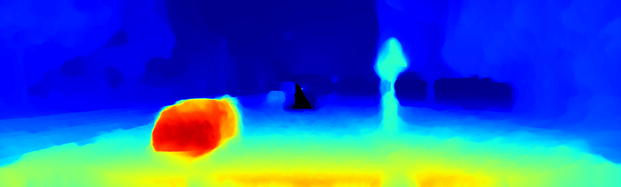}};
			\node[style={outer sep=0}, below left] at (pic.north east) {\small{\textcolor{white}{SFM}}};
		  \end{tikzpicture}\\
					 \begin{tikzpicture}[]
			\node[style={inner sep =2, outer sep=0}] (pic) at (0,0) {\includegraphics[width=0.23\textwidth]{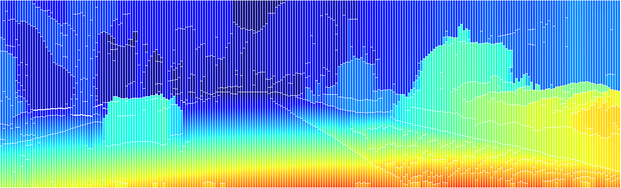}};
			\node[style={outer sep=0}, below left] at (pic.north east) {\small{\textcolor{white}{Mono-Stixels}}};
		  \end{tikzpicture}
		 & \begin{tikzpicture}[]
			\node[style={inner sep = 2, outer sep=0}] (pic) at (0,0) {\includegraphics[width=0.23\textwidth]{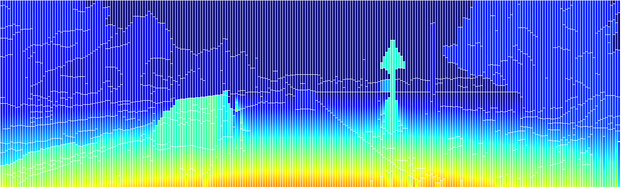}};
			\node[style={outer sep=0}, below left] at (pic.north east) {\small{\textcolor{white}{Mono-Stixels}}};
		  \end{tikzpicture} 
	\end{tabular}
  \caption{Depth representation obtained from structure-from-motion (SFM) and our proposed mono-stixel approach. While the SFM-based reconstruction fails for the preceding vehicle (left) and oncoming vehicle (right), the mono-stixels provide a reliable and compact representation of the whole dynamic scene. The color encodes the inverse depth from close (red) to far (dark blue).}
	\label{fig_intro}
\end{figure}
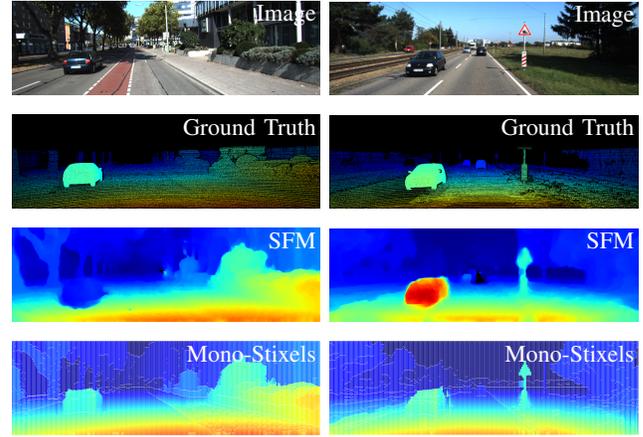

The mentioned works build on a dense disparity map from a stereo sensor. 
In general a disparity can be seen as a scaled value for the inverse depth which is also derivable from the optical flow induced by a single moving camera. 
However, this structure-from-motion principle does not hold for moving objects as shown in Fig. \ref{fig_intro} 
and thus the mentioned works are not applicable for a monocamera in dynamic scenes.

Therefore, this work introduces mono-stixels, a compact environment representation derived from a monocular camera sequence. 
Inputs for our methods are a dense optical flow field, pixel-wise semantic segmentation and camera motion estimation. 
Mono-stixels are especially designed to handle static and moving parts in a joint optimization resulting in a reliable depth reconstruction 
of the whole dynamic scene as shown in Fig. \ref{fig_intro}.

%% file: text/Introduction/RelatedWork.tex
We see three categories of related work. 
The first category consists of monocular depth estimates based on the structure-from-motion or multi-view geometry principle 
\cite{hartley2003multiple}. 
Mentionable approaches are PMO \cite{fananipmo}, DTAM \cite{newcombe2011dtam}, LSD-SLAM \cite{engel2014lsd}, ORB-SLAM \cite{mur2015orb} or REMODE \cite{pizzoli2014remode}. 
Based on the optical flow, the camera motion is estimated and the depth of the environment is derived based on the principle of multi-view geometry.
However, this principle only holds for static scenes and none of these methods can handle moving objects. 
Klappstein \cite{klappstein2008optical} proposed a way to detect moving objects based on the optical flow and camera motion. 
Static points in the scene have to fulfill the epipolar geometry, the positive depth as well as the positive height constraint.
But there are still some epipolar conformant independently moving objects (IMOs) like oncoming vehicles that are not detectable.
Ranftl et al. \cite{ranftl2016dense} distinguished objects in the scene by their fundamental matrix and proposed to reconstruct 
the points for each fundamental matrix individually. 
As a second step the different reconstructions are scaled in that way that they are connected, 
e.g. that a moving vehicle stands on the ground plane. 
However, epipolar conformant IMOs are represented by the same fundamental matrix as the static scene. 
Thus, these objects are not distinguishable and the reconstruction fails. 
Consequently, current structure-from-motion based approaches are limited to static scenes and IMOs that violate the epipolar constraint.

The second category comprises methods that provide a pixel-wise semantic segmentation or leverage this kind of information for a different vision task. 
The Cityscape dataset \cite{cordts2016cityscapes} gives an overview of the performance of pixel-wise semantic segmentation methods in street scenes 
and shows that deep neural networks allow to perform semantic segmentation on a previously unprecendented level of performance. 
One of the pioneering works are the fully convolutional networks (FCN) introduced by Long et al. \cite{long2015fully}. 
Semantics can be an useful information for different vision tasks. 
For example Bai et al. \cite{bai2016exploiting} use the semantic information to distinguish static objects from potentially moving traffic participants for optical flow estimation. 
Furthermore, Schneider et al. \cite{schneider2016semantic} showed that the semantic information can support the stereo-based stixel estimation resulting in a higher depth accuracy.

The last category of related works are the monocamera-based stixel estimation methods. 
Wolcott and Eustice \cite{wolcott2016probabilistic} proposed a column-wise partitioning of the image in ground, 
obstacle and background based on a prior appearance ground map and optical flow. 
Levi et al. \cite{levi2015stixelnet} introduced the Stixel-Net, a convolutional neural network for the detection of the ground contact point of the first obstacle in each column. 
However, both methods do not provide a multi-layer stixel world including a depth reconstruction of the whole scene. 
Current monocamera-based stixel estimation methods are more related to a free space estimation or road segmentation method.

There are two main contributions provided by the mono-stixels approach presented here. 
First, mono-stixels are a novel approach to estimate a multi-layer stixel world from a monocular camera sequence providing a depth reconstruction of the whole dynamic scene. 
Second, mono-stixels exploit semantic information and scene constraints for a joint monocular depth estimation of the static and moving parts of street scene. 
Thereby, mono-stixels provide reliable depth estimates even for the epipolar conformant IMOs, a novelty compared to previous structure-from-motion approaches.

%% file: text/Method/Method.tex
In this chapter, we describe our stixels estimation method. In the first section we define the mono-stixels model and segmentation problem which is formulated as an energy minimization problem in the second section. Finally, the last section describes the inference via dynamic programming. \par Our method assumes to have a dense optical flow field, a pixel-wise semantic segmentation and the camera motion as inputs. The dense optical flow field is defined as the motion of each image point from the current image to a previous image including a confidence measure. If the confidence is not provided by the optical flow algorithm itself, this could be estimated based on the structure tensor as described in \cite{Foerstner1984}. For the semantic information, we require that a pixel-wise semantic segmentation algorithm provides for each pixel the pseudo-probabilities (scores) to belong to the semantic classes listed in Table \ref{tab_stixeltypes}. 
\subsection{Mono-Stixels} 
Mono-stixels are defined as thin stick-like planar and rigid moving elements in the scene. To reduce the segmentation problem to a 1D optimization problem, the image of width $w$ and height $h$ is divided into columns of a fixed width $w_s$ and the segmentation problem is formulated and solved independently for each column as in \cite{pfeiffer2011multilayer}.
\begin{equation}
\begin{tabular}{c}
$\vs = \left\{s_i \mid 1 \leq i \leq N \leq h\right\}$ \\
with $s_i = (v_i^b, v_i^t, m_i, p_i, \vt_i, c_i)$
\end{tabular}
\label{eqn_segmentationproblem}
\end{equation}
The vector $\vs$ represents the segmentation of the column in $N$ mono-stixels, where each mono-stixel $s_i$ is defined by its bottom and top image coordinates $v_i^b, v_i^t$, its semantic class $c_i$, its stixel type $m_i$, its inverse depth $p_i$ and its motion $\vt_i$. Regarding the characteristic of traffic participants, the motion $\vt_i$ is defined as a 2D-translation of the stixel on the ground plane. The rotational motion is neglected due to the small horizontal extent of a stixel. \par 

\begin{table}[htpb]
\caption{Definition of mono-stixel types}
	\centering
	\begin{tabular}{p{0.13\textwidth} | p{0.11\textwidth} | p{0.07\textwidth} | p{0.08\textwidth}}
	Stixel type & Semantic classes & Geometry & Motion \\ \hline \hline
	ground $\sup$ & road, sidewalk, terrain & lying & static \\ \hline
	static object $\sobj$ & building, poles/signage, vegetation & upright & static \\ \hline
	dynamic object $\dobj$ & vehicle, \newline two-wheeler, \newline person & upright & potentially moving \\ \hline
	sky $\sky$ & sky & infinite \newline distance & static \\
	\end{tabular}	
	\label{tab_stixeltypes}
\end{table}

We define four stixel types that are solely distinguishable by their geometry and motion (see Table \ref{tab_stixeltypes}). Furthermore, we associate each semantic class to exactly one stixel type. Thereby, we leverage the semantic segmentation to prefer a specific stixel type. For example, a high score for the vehicle class prefers a dynamic object stixel.  

\subsection{Energy minimization problem} 
The segmentation problem of one column in Eq. (\ref{eqn_segmentationproblem}) is formulated as an energy minimization problem. The energy term captures a data likelihood $\Phi(\vs,\vf, \vl)$ depending on the optical flow $\vf$ and the pixel-wise semantic segmentation $\vl$. Additionally, a pairwise prior term $\Psi(\vs)$ incorporates prior knowledge about the typical structure of street scenes as in \cite{pfeiffer2011multilayer}.
\begin{eqnarray}
\hat{\vs}&{}={}&\min_{\vs}  E(\vs,\vf, \vl)  \nonumber \\
&{}={}&\min_{\vs} \left( \Psi(\vs) + \Phi(\vs,\vf, \vl) \right)
\label{eq_energyminimization}
\end{eqnarray}

The prior term prefers a stixel segmentation with a geometric layout likely for the typical structure of street scenes $\Psi_{str}(s_i,s_{i-1})$ and further regulates the model complexity by adding a constant value $\beta_{mc}$ for each new stixel.
\begin{equation}
\Psi(\vs) = \sum_{i=1}^{N} \left( \Psi_{str}(s_i,s_{i-1}) + \beta_{mc} \right)
\end{equation}
We follow the proposed structural prior term of Pfeiffer and Franke \cite{pfeiffer2011multilayer} in a slightly different definition. First, the gravity constraint prefers objects  standing on the ground plane. Thus, if an object stixel $s_{i} \in \{\sobj, \dobj \}$ succeeds a ground or sky stixel $s_{i-1} \in \{\sup, \sky \}$, the structural energy term is defined as $\Psi_{str}(s_i,s_{i-1}) = \Psi_{grav}$ with
\begin{equation}
\Psi_{grav} = \begin{cases}
\alpha_{grav}^{-} + \beta_{grav}^{-} \Delta h_{i}  & \text{if}\  \Delta h_{i} < 0 \\
\alpha_{grav}^{+} + \beta_{grav}^{+} \Delta h_{i}  & \text{if}\  \Delta h_{i} > 0 \\
\end{cases}
\label{eq_grav}
\end{equation}
where $\alpha_{grav}$ and $\beta_{grav}$ are tunable parameters to express the prior knowledge. $\Delta h_i$ is the height of the stixel $s_i$ over the reference ground, which is defined by the ground stixels in that column. If there is not any ground stixel, the reference ground is defined by the height of the camera mounting position on the vehicle. \par
However, if the previous object is also an object stixel $s_{i}, s_{i-1} \in \{\sobj, \dobj \}$ the bottom of the stixel might not be the bottom of the object due to occlusion or a depth discontinuity inside the object. Therefore, in this case the structural prior term is defined as the minimum of the gravity and an ordering constraint $\Psi_{str}(s_i,s_{i-1}) = \min (\Psi_{grav}, \Psi_{ord})$ with
\begin{equation}
\Psi_{ord} = \begin{cases}
\alpha_{ord} + \beta_{ord} \left( \frac{1}{p_{i}} - \frac{1}{p_{i-1}} \right)& \text{if}\ p_{i} > p_{i-1} \\
0 & \text{otherwise} \\
\end{cases}
\end{equation}
$\alpha_{ord}$ and $\beta_{ord}$ are again tunable parameters. \par
Furthermore, we prefer small discontinuities in the height of the ground plane, e.g. caused by a slanted street or a sidewalk. Thus, for ground stixels $s_{i} \in \sup$ we define $\Psi_{str}(s_i,s_{i-1}) = \Psi_{flat}$ with
\begin{equation}
\Psi_{flat} = \alpha_{flat} + \beta_{flat} \Delta h_{i}^2 
\end{equation}
where $\alpha_{flat}$ and $\beta_{flat}$ are tunable parameters and $\Delta h_{i}$ is the height difference between the ground stixel $s_i$ and reference ground as in Eq. (\ref{eq_grav}).\par
The unary term or data likelihood $\Phi(\vs,\vf, \vl)$ rates the consistency of an individual stixel hypotheses $s_i$ based on the semantic segmentation $\Phi_L(s_i,\vl_v)$ and optical flow field $\Phi_F(s_i,\vf_v,v)$. The data likelihoods are modeled to be independent across the pixels and therefore independent across the rows $v$ in that column.  
\begin{equation}
\Phi(\vs,\vf, \vl) = \sum_{i=1}^{N} \sum_{v=v_i^b}^{v_i^t} \left(\delta_L \Phi_L(s_i,\vl_v) + \delta_F \Phi_F(s_i,\vf_v,v) \right)
\end{equation}
where $\delta_L$ and $\delta_F$ weights the data likelihood of each part. \\
The data likelihood of the semantic segmentation $\Phi_L(s_i,\vl_v)$ prefers stixel hypothesis having a semantic class $c_i$ with high class scores $\vl_v(c_i)$ inside the stixel segment. 
\begin{equation}
\Phi_L(s_i,\vl_v) = \min \left(\alpha_{L}, -\log (\vl_v(c_i))\right)
\end{equation}
where $\alpha_{L}$ regards that the class score might be not reliable in some cases. \par
Note that stixels of a given type $m_i$ can only have one of the associated semantic classes defined in Table \ref{tab_stixeltypes}. Thus, a high class score prefers both the corresponding semantic class and the stixel type associated with this class. \par
Analogously, the term $\Phi_F(s_i,\vf_v,v)$ rates the consistency of the optical flow for an individual stixel hypothesis. Due to the definition of a stixel to be a planar part of the scene, the expected optical flow $\hat{\vectorsymbol{f}}_v$ of one pixel $\vx_v$ can be described by a homography $\mH_i$ for a given stixel hypothesis $s_i$ \cite{hartley2003multiple}:
\begin{equation}
\begin{tabular}{c}
$\hat{\vectorsymbol{f}}_v = \mH_i \vx_v - \vx_v$ \\
with $\mH_i = \mK \left( \mR_{cam} - p_i \vt_{cam,i} \vn_{i}^T \right) \mK^{-1}$
\label{eq_homography}
\end{tabular}
\end{equation}
The normal vector $\vn_i$ is defined by the geometric definition of the stixel type $m_i$ to be lying or upright (see Table \ref{tab_stixeltypes}) and the assumption that the stixel is facing the camera center. Furthermore, for static stixel types the rotation matrix $\mR_{cam}$ and translation vector $\vt_{cam,i}$ is solely defined by the camera motion. For sky stixels, there is the special case that the inverse depth is zero $p_i=0$ which simplifies the homography to $\mH_i = \mK \mR_{cam} \mK^{-1}$. \par
In contrast to that, for dynamic stixels also the translational motion $\vt_i$ of the stixel itself has to be regarded. However, the expected optical flow is still describable by a homography with the translational vector $\vt_{cam,i}$ defined as the relative translation between camera and stixel hypothesis $s_i$. Thereby, the homography serves as the common description of the optical flow for static and dynamic stixels in a monocular camera setup. \par
Pfeiffer and Franke \cite{pfeiffer2011multilayer} proposed to define the data likelihood for stereo depth measurements based on the difference between expected and measured disparity. Similar to this approach we define our data likelihood based on the residual flow $\vr_{i,v} =  \vf_v - \hat{\vectorsymbol{f}}_v = \vf_v - (\mH_i \vx_v - \vx_v)$. Therefore, we define a measurement model of the optical flow estimates as a Gaussian mixture model consisting of a normal distribution for inliers with covariance $\cov_{v}$ and a broad uniform distribution for outliers. Switching to the log-domain and sum up the constant parts in one parameter $\alpha_{F}$ the following energy term is derived:
\begin{equation}
\Phi_F(s_i,\vf_v,v) = \min \left(\alpha_{F}, \log \left(\det \cov_{v} \right)  + \frac{\vr_{i,v}^T \cov_{v}^{-1} \vr_{i,v}}{2} \right)
\end{equation}

\subsection{Solving the mono-stixels segmentation problem} 
The previous sections describe the mono-stixels segmentation problem (Eq. (\ref{eqn_segmentationproblem})) for one column as a 1D energy minimization problem (Eq. (\ref{eq_energyminimization})). This 1D energy minimization problem is solvable via dynamic programming, e.g. by using the Viterbi algorithm. However, even with dynamic programming the run time grows quadratic with all possible labels for a stixel hypothesis which results in a high computational effort. \par Therefore, we follow the proposed simplification to a minimum path problem in \cite{cordts2017stixel}. Only the stixel types $m_i$ and segmentation labels $v_i^b, v_i^t$ are optimized via dynamic programming while the other labels are approximated. Approximation in this context means to find the inverse depth $p_i$, semantic class $c_i$ and motion $\vt_i$ labels for one stixel hypothesis given its segmentation $v_i^b,v_i^t$ and type $m_i$. \par 
As the semantic class label $c_i$ we take that class associated with the stixel type $m_i$ that has the highest class scores in the corresponding image segment $[v_i^b, v_i^t]$.
\begin{equation}
c_i = \min_{c \in m_i} \sum_{v=v_i^b}^{v_i^t} \min \left(\alpha_{L}, -\log (\vl_v(c))\right)
\end{equation}

For the inverse depth $p_i$ and motion label $\vt_i$ in the first step we optimize the optical flow related energy term. That means we need to find the best homography $\mH_i$ for the corresponding image segment $[v_i^b, v_i^t]$ considering the geometry and motion definitions of the stixel type $m_i$ .
\begin{equation}
\mH_i = \min_{\mH \in m_i} \sum_{v=v_i^b}^{v_i^t} \Phi_F(\mH,\vf_v,v)
\label{inf_homography}
\end{equation}
In general, a homography has eight degrees of freedom. However, our stixel model allows to apply several constraints to reduce the degrees of freedom. $\mK$, $\mR_{cam}$ and $\vn_i$ are defined by the intrinsic calibration, camera motion or stixel type $m_i$ as discussed for Eq. (\ref{eq_homography}) and only the inverse depth $p_i$ and translation $\vt_{cam,i}$ are free parameters. Assuming a translational motion of the camera and stixel on the ground plane, the translation vector of the homography $\vt_{cam,i}$ can be described by two degrees of freedom. Thus, due to the linear dependency of the inverse depth $p_i$ and $\vt_{cam,i}$ in Eq. (\ref{eq_homography}) the homography $\mH$ has two degrees of freedoms, namely $\tilde{t}_x$ and $\tilde{t}_y$ with $p_i \vt_{cam,i} = \mR_{v2c} \matrix{\tilde{t}_x & \tilde{t}_y & 0}^T$ where $\mR_{v2c}$ rotates the 2D-translation on the ground plane in the camera coordinate system. Furthermore, for static object and ground stixels the translation is solely defined by the camera motion which means that only the inverse depth $p_i$ is left as one degree of freedom of the homography $\mH$. \par Exploiting the mono-stixel model shows that a stixel-homography only has up to two degrees of freedom for a given stixel type. This means that one optical flow vector $\vf_v$ is enough to derive a stixel-homography by solving the system $\mA p_i = \vb$ for static stixel and $\mA \matrix{\tilde{t}_x & \tilde{t}_y}^T = \vb$ for dynamic stixel. These systems are defined by rearranging the equation $\vf_v =  (\mH \vx_v - \vx_v)$.\par
To find the best homography $\mH_i$ for the whole image segment $[v_i^b, v_i^t]$ as defined in equation (\ref{inf_homography}), we use a MLESAC-based \cite{Torr2000} approach:
\begin{enumerate}
\item Take one optical flow vector $\vf_v \in \{\vf_{v_i^b},...,\vf_{v_i^t}\}$ to compute a hypothesis for the stixel-Homography $\mH$.
\item Compute the optical flow related energy term $\sum_{v=v_i^b}^{v_i^t} \Phi_F(\mH,\vf_v,v)$ for this homography
\item Repeat step one until all optical flow vectors are used 
\item Choose that homography $\mH$ with the lowest energy
\end{enumerate}

For static object and ground stixels, this directly defines the inverse depth label $p_i$ of the stixel as the degree of freedom of the homography $\mH_i$. \par
In contrast to that, for dynamic object stixel only the linear combination $p_i \vt_{cam,i}$ is defined by $\tilde{t}_x, \tilde{t}_y$ and either the inverse depth $p_i$ or one component of the translation vector $\vt_i$ can still be chosen freely. Based on our scene model it is possible to choose a plausible inverse depth $p_i$ by taking that one that minimizes the structural prior term $\Psi_{str}(s_i,s_{i-1})$. If the previous stixel $s_{i-1}$ is an ground stixel, this is achieved by that inverse depth that leads the stixel to stand on the ground stixel. If the previous stixel is an object stixel, the inverse depth is less clear as the ordering constraint is zero for every stixel behind the previous one. However, in this case we take the same depth as the previous stixel which might be roughly correct if both stixels belong to the same object. \par
Note that depending on the application it might not be required to choose a certain value for the inverse depth. The energy terms excluding the structural prior term are still defined by $\tilde{t}_x$ and $\tilde{t}_y$ and these values are able to represent the time to contact in the longitudinal and lateral direction with the camera as reference point. This might be enough for a criticality analysis. \par
Based on this approximations all labels are defined during the inference via dynamic programming. The achieved run time of this optimization is $\Onot{h^3}$ for one column and thereby $\Onot{\frac{w}{w_s} h^3}$ for all columns for an image of width $w$ and height $h$.

%% file: text/Experiments/Experiments.tex
\begin{figure*}[th]
\vspace{0.2cm}
  \centering
  \hspace*{-0.3cm}\begin{tabular}{@{} p{0cm}@{\hskip 0.02in} c@{\hskip 0.02in} c@{\hskip 0.02in} c@{\hskip 0.02in} c@{\hskip 0.02in} @{}}
	\multicolumn{1}{c}{\RotText{\scriptsize{Image}}} & \includegraphics[width=0.232\textwidth]{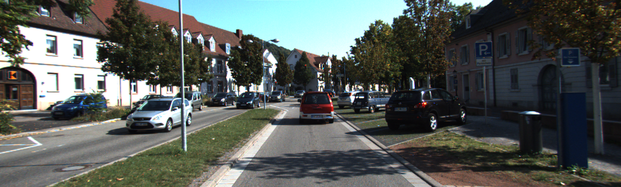}  & \includegraphics[width=0.232\textwidth]{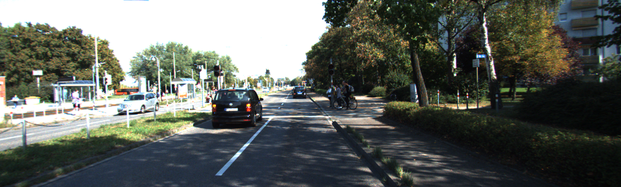} & \includegraphics[width=0.232\textwidth]{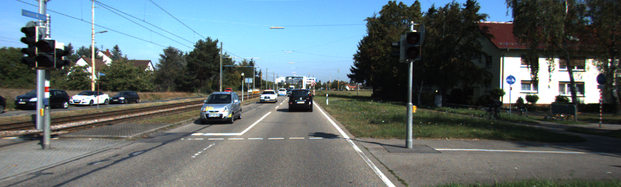} & \includegraphics[width=0.232\textwidth]{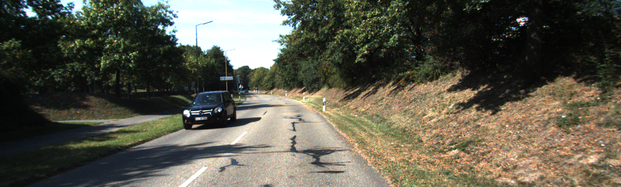} \\
	\multicolumn{1}{c}{\RotText{\scriptsize{Input\\Flow\cite{weinzaepfel2013deepflow}}}} & \includegraphics[width=0.232\textwidth]{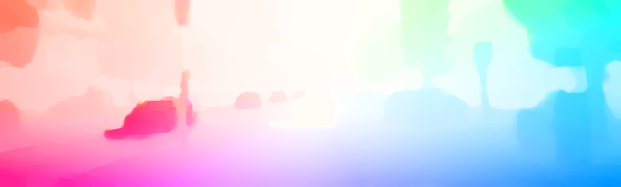}  & \includegraphics[width=0.232\textwidth]{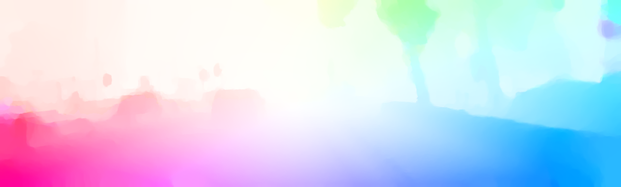} & \includegraphics[width=0.232\textwidth]{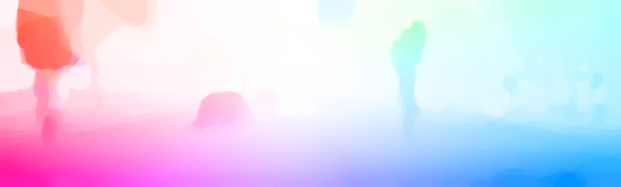} & \includegraphics[width=0.232\textwidth]{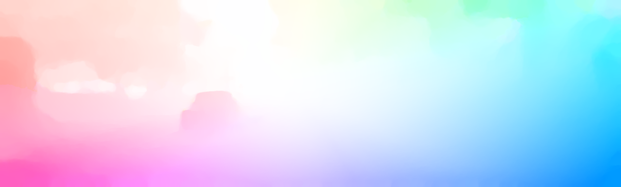} \\
	\multicolumn{1}{c}{\RotText{\scriptsize{Input\\Semantic\cite{long2015fully}}}} & \includegraphics[width=0.232\textwidth]{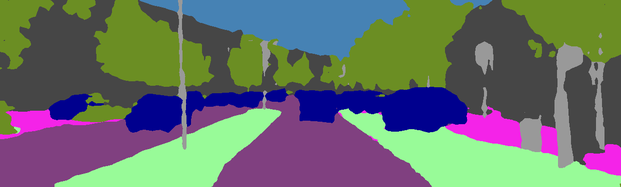}  & \includegraphics[width=0.232\textwidth]{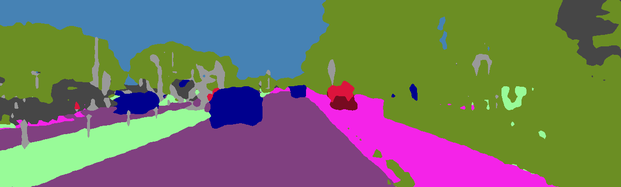} & \includegraphics[width=0.232\textwidth]{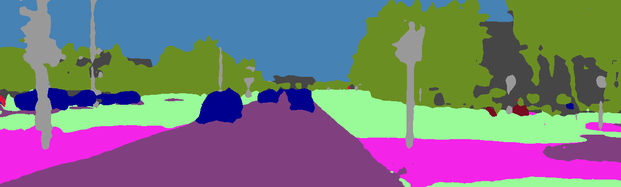} & \includegraphics[width=0.232\textwidth]{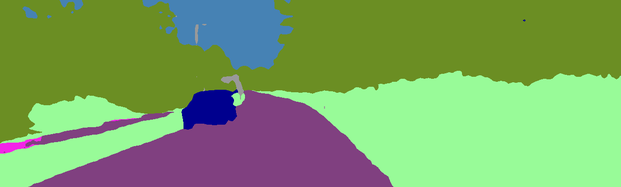}\\
	\multicolumn{1}{c}{\RotText{\scriptsize{Mono-Stixel\\Depth}}}  & \includegraphics[width=0.232\textwidth]{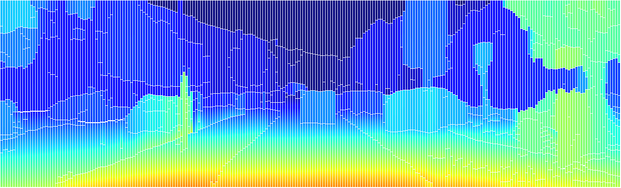}  & \includegraphics[width=0.232\textwidth]{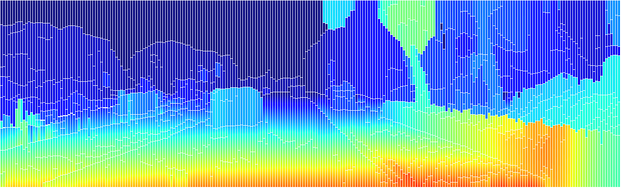} & \includegraphics[width=0.232\textwidth]{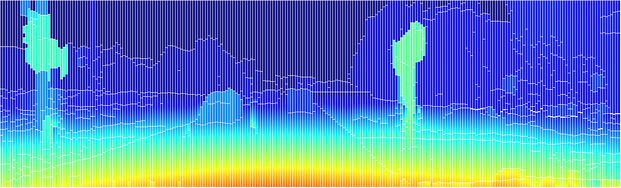} & \includegraphics[width=0.232\textwidth]{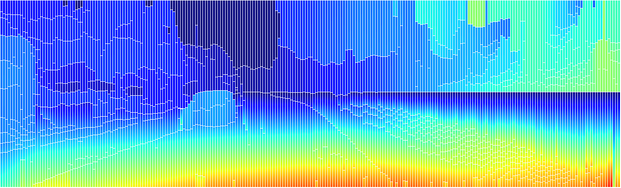}\\
	\multicolumn{1}{c}{\RotText{\scriptsize{Mono-Stixel\\Semantic}}} & \includegraphics[width=0.232\textwidth]{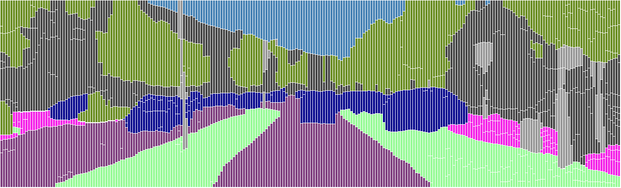}  & \includegraphics[width=0.232\textwidth]{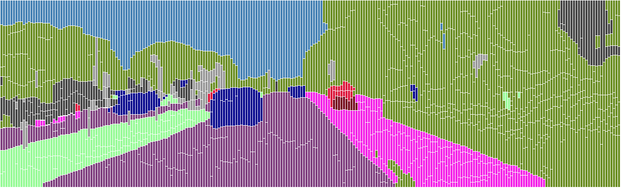} & \includegraphics[width=0.232\textwidth]{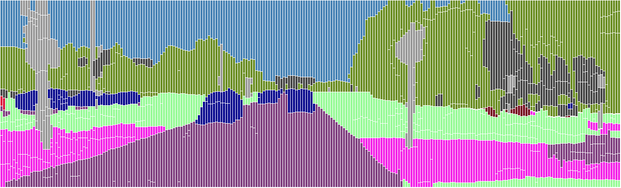} & \includegraphics[width=0.232\textwidth]{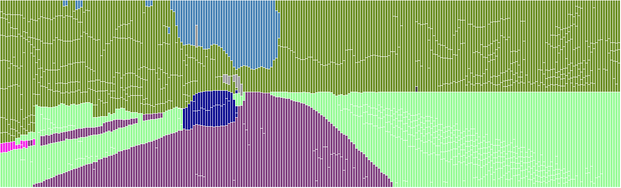}\\
	\end{tabular}
  \caption{Example depth and semantic representation of the mono-stixels based on the optical flow and semantic segmentation. The color encodes the inverse depth from close (red) to far (dark blue) or the semantic class following \cite{cordts2016cityscapes}. }
	\label{fig_exampleoutput}
\end{figure*}

In this chapter we give an in-depth analysis of the performance of our proposed mono-stixels. First, we describe the setup of our experiments comprising the used dataset and inputs. Second, we define a metric and baseline and finally we present an evaluation based on this metric and show some example results of our proposed mono-stixels.
\subsection{Setup}
We evaluate our approach on the KITTI-Stereo'15 \cite{Menze2015CVPR} dataset that contains 200 images captured from a forward facing camera in different street scenes. The dataset provides sparse ground truth depth obtained from a Velodyne HDL-64 laser scanner and 3D-CAD models for moving vehicles. Furthermore, based on the scene flow static and moving objects are distinguishable which enables us a more distinctive evaluation. \par
As mentioned, our method assumes a dense optical flow field, camera motion and a pixel-wise semantic segmentation as inputs. In our experiments the camera motion is provided by the visual odometry method in \cite{geiger2011stereoscan}. The dense optical flow field is estimated by the public available DeepFlow \cite{weinzaepfel2013deepflow}. Referring to the KITTI optical flow benchmark \cite{Menze2015CVPR}, there are also real-time capable dense optical flow methods with comparable or even better performance than DeepFlow. The optical flow is estimated on keyframes with a minimum driven distance of $0.5 m$. These keyframes do not exist for all images in the dataset, thus, the evaluation is done on 171 images out of the 200 images of the KITTI-Stereo'15 dataset. \par
For the pixel-wise semantic segmentation we train our own FCN8s network \cite{long2015fully} based on the VGG architecture \cite{simonyan2014very}. We follow the proposed training in \cite{schneider2016semantic}. First, we train our net on the Cityscape dataset \cite{cordts2016cityscapes}. Second, we fine-tune this net on 470 images of the KITTI dataset collected from the labeled subsets in \cite{kundu2014joint}, \cite{ros2015vision}, \cite{sengupta2013urban}, \cite{xu2016multimodal}, \cite{ovsep2016multi}, \cite{upcroft2014lighting}. Furthermore, we unified the training data to the 10 classes mentioned in Table \ref{tab_stixeltypes}. 
\subsection{Metric and Baseline}
To evaluate our method we follow the proposed metric in \cite{eigen2014depth}. $T$ is defined as the set of ground truth values $y_i^*$ and $y_i$ is the corresponding estimated depth at that position.
\begin{table}[h]
	\centering
	\begin{tabular}{ l  m{0.26\textwidth} r} 
	RMSE & \centering $\sqrt{ \frac{1}{|T|} \sum_{i \in T} ||y_i - y_i^* ||^2 }$ & (lower is better) \\ 
	Rel. Error & \centering $\frac{1}{|T|} \sum_{i \in T} \frac{|y_i - y_i^* |}{y_i^*}$ & (lower is better) \\ 
	Threshold & \centering \% of $i \in T$ s.t. $\max \left(\frac{y_i}{y_i^*}, \frac{y_i^*}{y_i} \right) = \delta < thr$ & (higher is better)
	\end{tabular} 
\end{table}

Additionally, we indicate the compactness by the number of values needed to represent the depth of the environment. Three values per mono-stixel are needed to describe the depth of the scene: one for the segmentation, one for the stixel type and one for the inverse depth. Thus, the compactness is defined as three times the mean number of mono-stixels per image. For a dense depth map the compactness is equal to the number of pixel per image. \par
As described in the related work section, the structure-from-motion approaches are still limited to static scenes or IMOs that violate the epipolar constraint. Our experiments should clearly show that our methods provide a reliable depth reconstruction of moving objects. Therefore, we separately evaluate the static and moving parts of the scene. Note, that many objects in street scene are epipolar conformant IMOs and it is essential to handle these objects to produce good results. Furthermore, we implement as a baseline a traditional structure-from-motion method by performing a triangulation \cite{hartley2003multiple} for each image point based on a dense optical flow field and camera motion. As inputs of this method we use exactly the same dense optical flow and camera motion estimation as for our mono-stixels. Thus, the comparison to this baseline directly shows the impact of the mono-stixels and is not effected by the performance of the used optical flow or camera motion estimation methods.
\subsection{Results}
\begin{figure*}[htpb]
\vspace{0.2cm}
  \centering
  \begin{tabular}{@{}c@{\hskip 0.03in} c@{\hskip 0.03in} c@{\hskip 0.03in} c@{\hskip 0.03in} @{}}
	Image & Ground Truth & Baseline - SFM \cite{hartley2003multiple}& Ours - Mono-Stixels \\
	\includegraphics[width=0.24\textwidth]{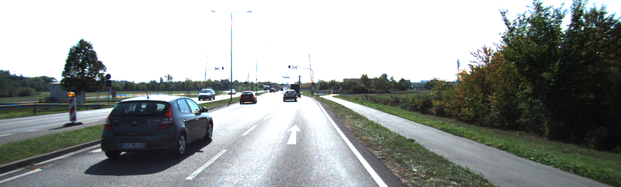} & \includegraphics[width=0.24\textwidth]{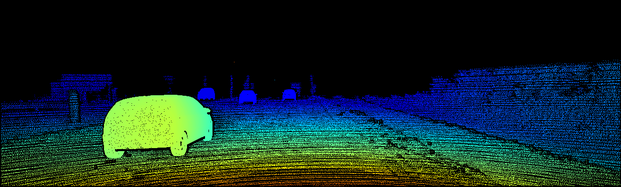} & \includegraphics[width=0.24\textwidth]{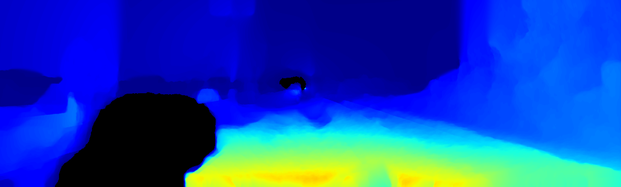} &  \includegraphics[width=0.24\textwidth]{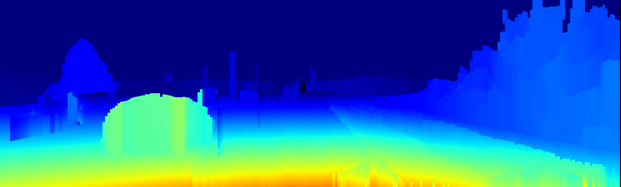}\\
	\includegraphics[width=0.24\textwidth]{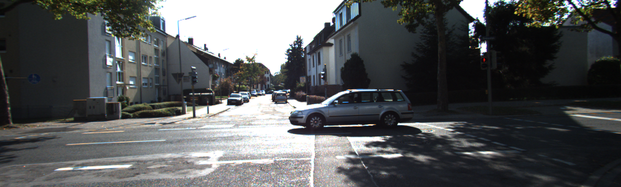} & \includegraphics[width=0.24\textwidth]{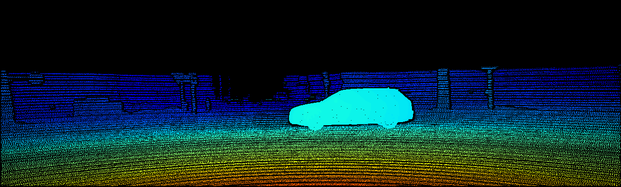} & \includegraphics[width=0.24\textwidth]{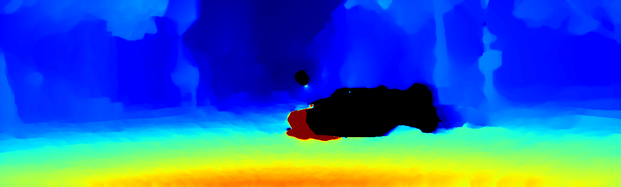} &  \includegraphics[width=0.24\textwidth]{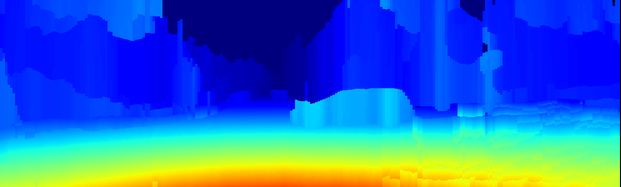}\\
	\includegraphics[width=0.24\textwidth]{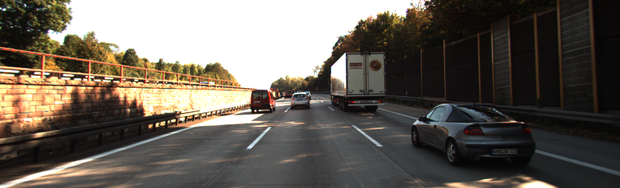} & \includegraphics[width=0.24\textwidth]{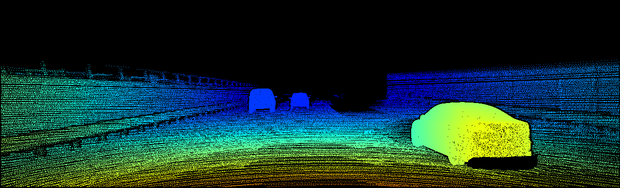} & \includegraphics[width=0.24\textwidth]{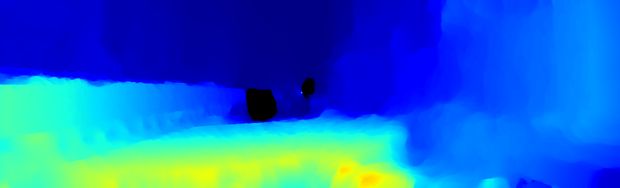} &  \includegraphics[width=0.24\textwidth]{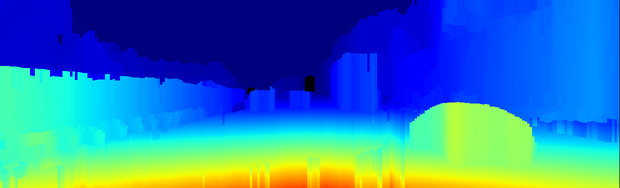}\\
	\includegraphics[width=0.24\textwidth]{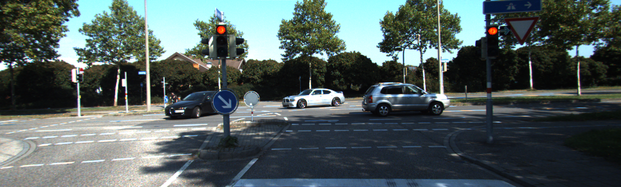} & \includegraphics[width=0.24\textwidth]{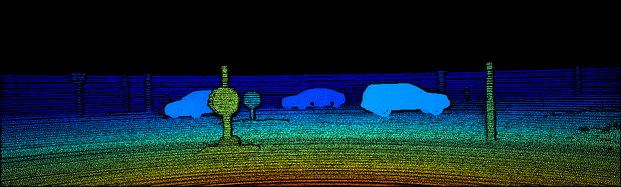} & \includegraphics[width=0.24\textwidth]{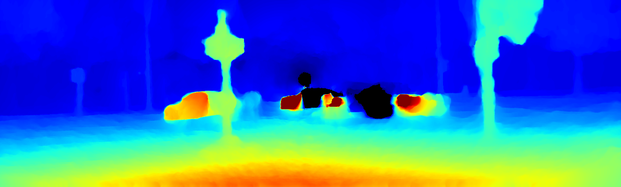} &  \includegraphics[width=0.24\textwidth]{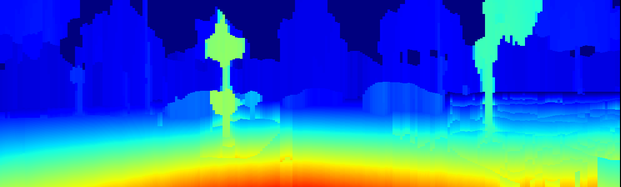}\\
	\includegraphics[width=0.24\textwidth]{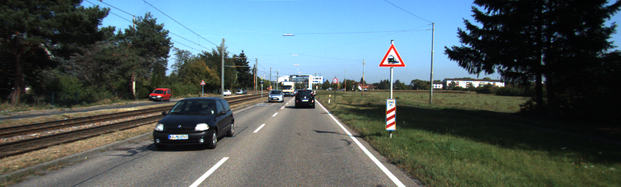} & \includegraphics[width=0.24\textwidth]{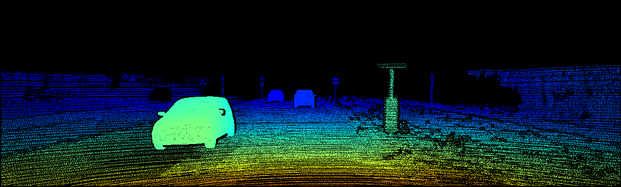} & \includegraphics[width=0.24\textwidth]{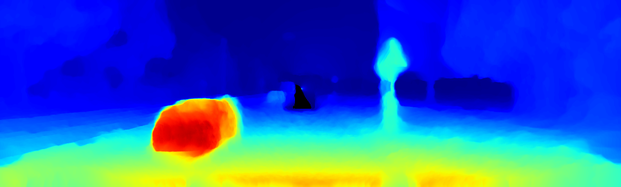} &  \includegraphics[width=0.24\textwidth]{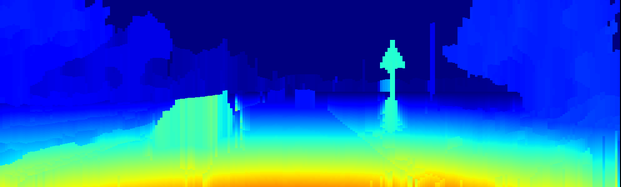}\\
	\end{tabular}
  \caption{Example performance of depth reconstruction on KITTI-Stereo'15 compared to a structure-from-motion (SFM) approach \cite{hartley2003multiple}. The color encodes the inverse depth from close (red) to far (dark blue). Invalid negative depth values are colored black.}
	\label{fig_exampleperformance}
\end{figure*}
\begin{table}[thpb]
\caption{Results of mono-stixels on KITTI-Stereo'15 compared to a structure-from-motion (Sfm) method }
\centering
\renewcommand{\arraystretch}{1.2}
\begin{tabular}{c c | c | c}
&  & Baseline -  & Ours -  \\
& Metric & SFM \cite{hartley2003multiple} & Mono-Stixels \\ \hline \hline
& Compactness & 465 k & \textbf{5.69 k} \\ \hline
\multirow{6}{*}{\rotatebox[origin=c]{90}{Overall}} & Rel. Error [\%] & 35.57 & \textbf{12.30} \\
& RMSE [m] & 11.29 & \textbf{6.29}\\
 & $\delta<1.1$ [\%]& 66.59 & \textbf{70.61}\\
& $\delta<1.25$ [\%]& 76.98 & \textbf{88.95}\\
 & $\delta<1.25^2$ [\%]& 82.90 & \textbf{95.83}\\
& $\delta<1.25^3$ [\%]& 87.44 & \textbf{97.73}\\ \hline
\multirow{6}{*}{\rotatebox[origin=c]{90}{Static}} & Rel. Error [\%] & 17.08 & \textbf{11.22}\\
& RMSE [m]& 7.93 & \textbf{6.31}\\
 & $\delta<1.1$ [\%]& \textbf{76.18} & 76.03\\
& $\delta<1.25$ [\%]& 87.57 & \textbf{90.42}\\
 & $\delta<1.25^2$ [\%]& 93.19 & \textbf{95.95}\\
& $\delta<1.25^3$ [\%]& 95.65 & \textbf{97.73}\\ \hline
\multirow{6}{*}{\rotatebox[origin=c]{90}{Moving}} & Rel. Error [\%] & 158.07 & \textbf{19.40}\\
& RMSE [m] & 23.56 & \textbf{6.16}\\
 & $\delta<1.1$ [\%]& 3.16 & \textbf{34.69}\\
& $\delta<1.25$ [\%]& 6.94 & \textbf{79.20}\\
 & $\delta<1.25^2$ [\%]& 14.75 & \textbf{95.04}\\
& $\delta<1.25^3$ [\%]& 33.10 & \textbf{97.72}
\end{tabular}
\label{tab_results}
\end{table}

Fig. \ref{fig_exampleoutput} shows some example outputs of the proposed mono-stixels using a stixel width of $w_s = 5$ and hand-tuned parameters. Please note that we additionally provide a video consisting of a 3D visualization in the supplementary material. \par
These examples show that mono-stixels provide a compact and plausible reconstruction of the static as well as the dynamic parts of the scene. 
The performance of mono-stixels compared to the structure-from-motion baseline is shown in Fig. \ref{fig_exampleperformance} including different moving objects like oncoming, preceding or crossing vehicles. The examples show that the structure-from-motion baseline fails for all moving objects while the mono-stixels provide a reliable depth reconstruction for all moving objects. This is also shown in the qualitative evaluation in Table \ref{tab_results}. Mono-stixels show a comparable and even more robust performance for the static parts compared to the baseline. But, only the mono-stixels are able to provide also reliable depth estimates for the moving objects. Furthermore, only one eightieth of the values is needed to represent the depth of the scene.\par
But we also show failure cases in the examples in Fig. \ref{fig_exampleoutput}. First, the mono-stixels depend on the performance of the used optical flow algorithm. In the first example, the optical flow estimation fails for the upper part of the pole on the left side which results in wrong depth estimates of that part. Furthermore, the last example shows that the estimation might fail for scenes that violate the world assumptions. In that example the high slope of the grass in the right part violate the assumption of a flat ground plane which results in a high number of segments in the lower part and wrong depth estimates of the upper part. Furthermore, a slanted ground plane would violate our assumption regarding the translational motion of a moving object. However, this should be solvable by applying the concept of slanted stixels \cite{hernandez2017slanted} that are specially designed to represent high slope in the ground plane.

%% file: text/Conclusion/Conclusion.tex
This paper presented mono-stixels, a novel stixel estimation method from monocular camera sequences. The homography-based formulation allows to describe the optical flow for static and dynamic stixels in a common way for a joint optimization. Furthermore, we showed how to leverage a pixel-wise semantic label to distinguish static and potentially moving objects and how to use a scene and ground plane model especially for the depth reconstruction of moving objects in this stixel estimation method. \par
Many works still showed the suitability of the medium-level-representation with stixels as primitive elements for high-level vision tasks. Thus, the mono-stixels approach could be the enabler to use the stixel world in a monocamera setup for driver assistance systems or autonomous vehicles. 